\pdfoutput=1

\documentclass[11pt]{article}

\usepackage[]{EACL2023}

\usepackage{times}
\usepackage{epsfig}
\usepackage{graphicx}
\usepackage{amsmath}
\usepackage{amssymb}
\usepackage{booktabs}
\usepackage{hyperref}
\usepackage{tabularx}
\usepackage{soul}
\usepackage{epstopdf}
\usepackage{todonotes}
\usepackage[utf8]{inputenc}
\usepackage{caption}
\usepackage{xstring}
\usepackage{color}
\usepackage{makecell}
\usepackage{latexsym}
\usepackage{booktabs}
\usepackage{pifont}
\usepackage{url}
\usepackage{multirow}
\usepackage{xspace}
\usepackage{subcaption}
\usepackage{endnotes}
\usepackage{array}
\usepackage{float}
\usepackage{mfirstuc, tabulary}
\usepackage{paralist}
\usepackage{float}
\usepackage{csquotes}
\usepackage{xcolor,colortbl}
\usepackage{svg}

\usepackage[T1]{fontenc}

\usepackage[utf8]{inputenc}

\usepackage{microtype}

\usepackage{inconsolata}

\title{Text-Derived Knowledge Helps Vision: A Simple Cross-modal Distillation for Video-based Action Anticipation}

\author{Sayontan Ghosh$^{1}$
        \And  
        Tanvi Aggarwal$^{1}$
        \AND
        Minh Hoai$^{1}$ 
        \And
        Niranjan Balasubramanian$^{1}$  \\
        \AND
        $^{1}$Stony Brook University
        \AND
        \{sagghosh, taggarwal, minhhoai, niranjan \}@cs.stonybrook.edu}

\begin{document}
\maketitle

\def\mA{\mathcal{A}}
\def\mB{\mathcal{B}}
\def\mC{\mathcal{C}}
\def\mD{\mathcal{D}}
\def\mE{\mathcal{E}}
\def\mF{\mathcal{F}}
\def\mG{\mathcal{G}}
\def\mH{\mathcal{H}}
\def\mI{\mathcal{I}}
\def\mJ{\mathcal{J}}
\def\mK{\mathcal{K}}
\def\mL{\mathcal{L}}
\def\mM{\mathcal{M}}
\def\mN{\mathcal{N}}
\def\mO{\mathcal{O}}
\def\mP{\mathcal{P}}
\def\mQ{\mathcal{Q}}
\def\mR{\mathcal{R}}
\def\mS{\mathcal{S}}
\def\mT{\mathcal{T}}
\def\mU{\mathcal{U}}
\def\mV{\mathcal{V}}
\def\mW{\mathcal{W}}
\def\mX{\mathcal{X}}
\def\mY{\mathcal{Y}}
\def\mZ{\mathcal{Z}} 

\def\bbN{\mathbb{N}} 
\def\bbR{\mathbb{R}} 
\def\bbP{\mathbb{P}} 
\def\bbQ{\mathbb{Q}} 
\def\bbE{\mathbb{E}}

\def\1n{\mathbf{1}_n}
\def\0{\mathbf{0}}
\def\1{\mathbf{1}}

\def\A{{\bf A}}
\def\B{{\bf B}}
\def\C{{\bf C}}
\def\D{{\bf D}}
\def\E{{\bf E}}
\def\F{{\bf F}}
\def\G{{\bf G}}
\def\H{{\bf H}}
\def\I{{\bf I}}
\def\J{{\bf J}}
\def\K{{\bf K}}
\def\L{{\bf L}}
\def\M{{\bf M}}
\def\N{{\bf N}}
\def\O{{\bf O}}
\def\P{{\bf P}}
\def\Q{{\bf Q}}
\def\R{{\bf R}}
\def\S{{\bf S}}
\def\T{{\bf T}}
\def\U{{\bf U}}
\def\V{{\bf V}}
\def\W{{\bf W}}
\def\X{{\bf X}}
\def\Y{{\bf Y}}
\def\Z{{\bf Z}}

\def\a{{\bf a}}
\def\b{{\bf b}}
\def\c{{\bf c}}
\def\d{{\bf d}}
\def\e{{\bf e}}
\def\f{{\bf f}}
\def\g{{\bf g}}
\def\h{{\bf h}}
\def\i{{\bf i}}
\def\j{{\bf j}}
\def\k{{\bf k}}
\def\l{{\bf l}}
\def\m{{\bf m}}
\def\n{{\bf n}}
\def\o{{\bf o}}
\def\p{{\bf p}}
\def\q{{\bf q}}
\def\r{{\bf r}}
\def\s{{\bf s}}
\def\t{{\bf t}}
\def\u{{\bf u}}
\def\v{{\bf v}}
\def\w{{\bf w}}
\def\x{{\bf x}}
\def\y{{\bf y}}
\def\z{{\bf z}}

\def\balpha{\mbox{\boldmath{$\alpha$}}}
\def\bbeta{\mbox{\boldmath{$\beta$}}}
\def\bdelta{\mbox{\boldmath{$\delta$}}}
\def\bgamma{\mbox{\boldmath{$\gamma$}}}
\def\blambda{\mbox{\boldmath{$\lambda$}}}
\def\bsigma{\mbox{\boldmath{$\sigma$}}}
\def\btheta{\mbox{\boldmath{$\theta$}}}
\def\bomega{\mbox{\boldmath{$\omega$}}}
\def\bxi{\mbox{\boldmath{$\xi$}}}
\def\bnu{\mbox{\boldmath{$\nu$}}}                                  
\def\bphi{\mbox{\boldmath{$\phi$}}}
\def\bmu{\mbox{\boldmath{$\mu$}}}

\def\bDelta{\mbox{\boldmath{$\Delta$}}}
\def\bOmega{\mbox{\boldmath{$\Omega$}}}
\def\bPhi{\mbox{\boldmath{$\Phi$}}}
\def\bLambda{\mbox{\boldmath{$\Lambda$}}}
\def\bSigma{\mbox{\boldmath{$\Sigma$}}}
\def\bGamma{\mbox{\boldmath{$\Gamma$}}}
                                  
\newcommand{\myprob}[1]{\mathop{\mathbb{P}}_{#1}}

\newcommand{\myexp}[1]{\mathop{\mathbb{E}}_{#1}}

\newcommand{\mydelta}[1]{1_{#1}}

\newcommand{\myminimum}[1]{\mathop{\textrm{minimum}}_{#1}}
\newcommand{\mymaximum}[1]{\mathop{\textrm{maximum}}_{#1}}    
\newcommand{\mymin}[1]{\mathop{\textrm{minimize}}_{#1}}
\newcommand{\mymax}[1]{\mathop{\textrm{maximize}}_{#1}}
\newcommand{\mymins}[1]{\mathop{\textrm{min.}}_{#1}}
\newcommand{\mymaxs}[1]{\mathop{\textrm{max.}}_{#1}}  
\newcommand{\myargmin}[1]{\mathop{\textrm{argmin}}_{#1}} 
\newcommand{\myargmax}[1]{\mathop{\textrm{argmax}}_{#1}} 
\newcommand{\myst}{\textrm{s.t. }}

\newcommand{\denselist}{\itemsep -1pt}
\newcommand{\sparselist}{\itemsep 1pt}

\definecolor{pink}{rgb}{0.9,0.5,0.5}
\definecolor{purple}{rgb}{0.5, 0.4, 0.8}   
\definecolor{gray}{rgb}{0.3, 0.3, 0.3}
\definecolor{mygreen}{rgb}{0.2, 0.6, 0.2}

\newcommand{\cyan}[1]{\textcolor{cyan}{#1}}
\newcommand{\red}[1]{\textcolor{red}{#1}}  
\newcommand{\blue}[1]{\textcolor{blue}{#1}}
\newcommand{\magenta}[1]{\textcolor{magenta}{#1}}
\newcommand{\pink}[1]{\textcolor{pink}{#1}}
\newcommand{\green}[1]{\textcolor{green}{#1}} 
\newcommand{\gray}[1]{\textcolor{gray}{#1}}    
\newcommand{\mygreen}[1]{\textcolor{mygreen}{#1}}    
\newcommand{\purple}[1]{\textcolor{purple}{#1}}       

\definecolor{greena}{rgb}{0.4, 0.5, 0.1}
\newcommand{\greena}[1]{\textcolor{greena}{#1}}

\definecolor{bluea}{rgb}{0, 0.4, 0.6}
\newcommand{\bluea}[1]{\textcolor{bluea}{#1}}
\definecolor{reda}{rgb}{0.6, 0.2, 0.1}
\newcommand{\reda}[1]{\textcolor{reda}{#1}}

\def\changemargin#1#2{\list{}{\rightmargin#2\leftmargin#1}\item[]}
\let\endchangemargin=\endlist
                                               
\newcommand{\cm}[1]{}

\newcommand{\mhoai}[1]{{\color{purple}\textbf{[MH: #1]}}}

\newcommand{\mtodo}[1]{{\color{red}$\blacksquare$\textbf{[TODO: #1]}}}
\newcommand{\myheading}[1]{\vspace{1ex}\noindent \textbf{#1}}
\newcommand{\htimesw}[2]{\mbox{$#1$$\times$$#2$}}


\newif\ifshowsolution
\showsolutiontrue

\ifshowsolution  
\newcommand{\Comment}[1]{\paragraph{\bf $\bigstar $ COMMENT:} {\sf #1} \bigskip}
\newcommand{\Solution}[2]{\paragraph{\bf $\bigstar $ SOLUTION:} {\sf #2} }
\newcommand{\Mistake}[2]{\paragraph{\bf $\blacksquare$ COMMON MISTAKE #1:} {\sf #2} \bigskip}
\else
\newcommand{\Solution}[2]{\vspace{#1}}
\fi

\newcommand{\truefalse}{
\begin{enumerate}
	\item True
	\item False
\end{enumerate}
}

\newcommand{\yesno}{
\begin{enumerate}
	\item Yes
	\item No
\end{enumerate}
}

\newcommand{\Sref}[1]{Sec.~\ref{#1}}
\newcommand{\Eref}[1]{Eq.~(\ref{#1})}
\newcommand{\Fref}[1]{Fig.~\ref{#1}}
\newcommand{\Tref}[1]{Table~\ref{#1}}

\begin{abstract}
Anticipating future actions in a video is useful for many autonomous and assistive technologies.
Most prior action anticipation work treat this as a vision modality problem, where the models learn the task information primarily from the video features in the action anticipation datasets.
However, knowledge about action sequences can also be obtained from external textual data. 
In this work, we show how knowledge in pretrained language models can be adapted and distilled into vision-based action anticipation models.
We show that a simple distillation technique can achieve effective knowledge transfer and provide consistent gains on a strong vision model (Anticipative Vision Transformer) for two action anticipation datasets (3.5\% relative gain on \texttt{EGTEA-GAZE+} and 7.2\% relative gain on \texttt{EPIC-KITCHEN 55}), giving a new state-of-the-art result\footnote{The models and code used are available at:\href{https://github.com/StonyBrookNLP/action-anticipation-lmtovideo}{https://github.com/StonyBrookNLP/action-anticipation-lmtovideo}}. 
\end{abstract}
\section{Introduction}\label{sec_intro}
Anticipating future actions in the video of an unfolding scenario is an important capability for many applications in augmented reality~\cite{ar_cite1, ar_cite2}, robotics~\cite{robotics1, robotics2}, and autonomous driving~\cite{self_drive1, self_drive2}. Anticipating what actions will likely happen in a scenario, requires one to both recognize what has happened so far, and use anticipative general knowledge about how action sequences tend to play out. Most models for this task use a pre-trained video encoder to extract information about what has happened so far in the scenario, and use a text-based decoder to predict what action is likely to happen in the future~\cite{carion2020end, dessalene2021forecasting, liu2020forecasting, sener2020temporal}.

However, when trained on the target video datasets, the generalization of the models depends on how well these video datasets cover the space of action sequence distributions. In other words, the knowledge that is learnt for predicting future actions is, in effect, limited to the information in the target video datasets, where obtaining large scale coverage of action sequences is difficult.

Knowledge about action sequences can also be obtained from text resources at scale.
Language models, (e.g. \texttt{BERT}~\cite{bert}, \texttt{RoBERTa}~\cite{roberta}), are typically pre-trained on large collections of unlabeled texts with billions of tokens, where they acquire a wide-variety of knowledge including large scale knowledge about common action sequences. For example, Table~\ref{tab:mlm_ex} illustrates how the pre-trained \texttt{BERT} is able to predict the next action in a sequence of actions extracted from a recipe video in terms of its verb and the object.
\begin{figure}
    \centering
    \includegraphics[width=0.95\linewidth]{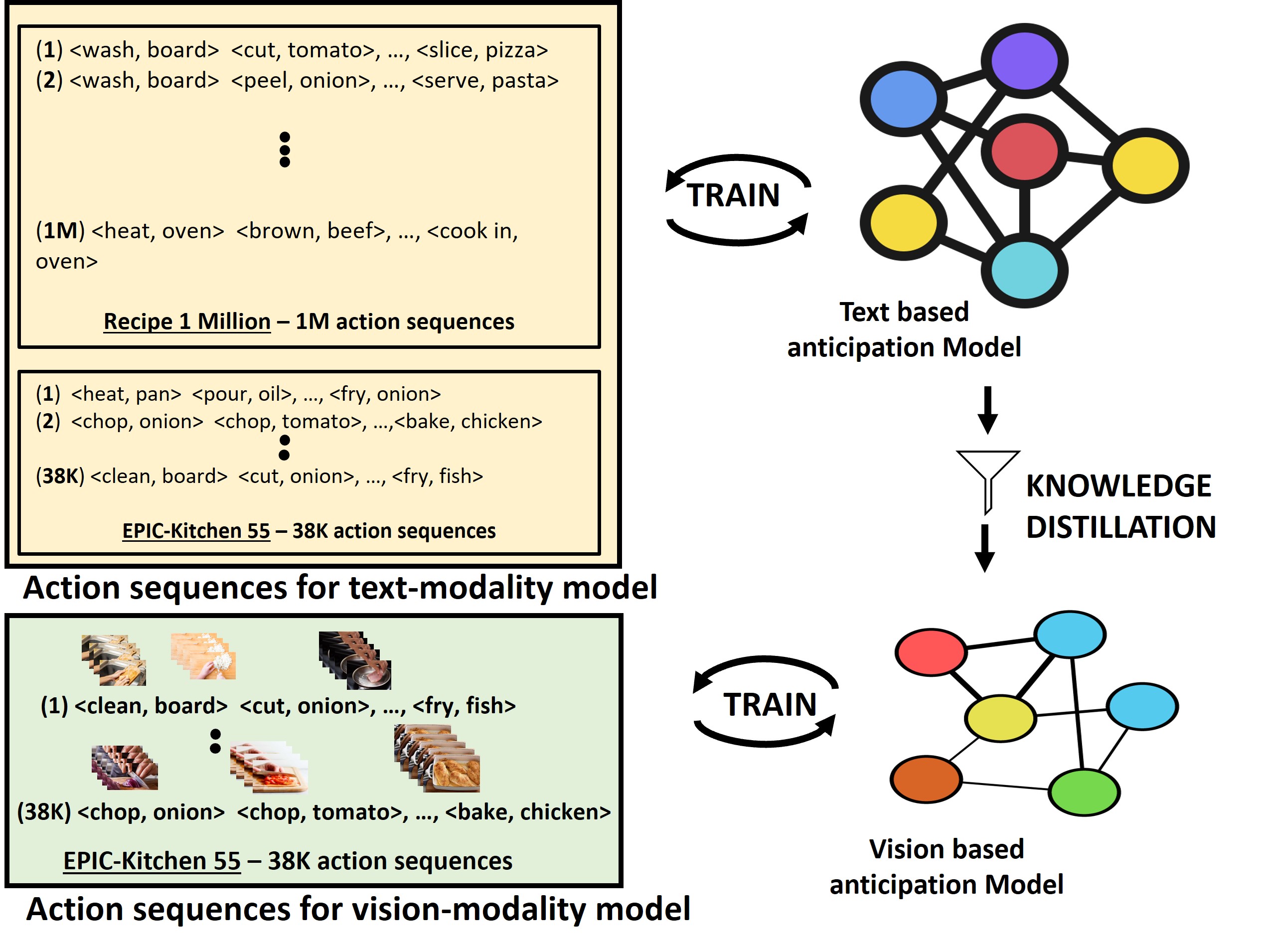}
    \caption{A model learning the action anticipation from only the vision modality (video frames) is essentially exposed to a very limited set of action sequences. Language models, which are pre-trained on large-scale text, can learn this distribution from the task, and a much larger domain-relevant text. We propose distilling this knowledge from text modality models to vision modality model for video action anticipation task.
    }
    \label{fig:main_idea}
\end{figure}
Also, it is easier to collect a much larger collection of action sequences from text sources compared to video annotated with segments. As illustrated in Figure~\ref{fig:main_idea}, \texttt{EPIC55}, a video dataset of about $800$\texttt{GB} only has about $38K$ action sequences, whereas there are around $1M$ sequences in the text recipes dataset \texttt{Recipe1M}. Text modality models can thus be exposed to a much larger variety of action sequences compared to video-modality anticipation models. However, because the task is defined only over the video inputs there is a question of how one can transfer this knowledge.

In this work, we show that we can augment video-based anticipation models with this external text-derived knowledge.
\begin{table}[t]
    \small
    \centering
    \begin{tabular}{ | m{15em} | m{3em}|  }
      \hline
      \textbf{Masked action sequence} & \textbf{BERT @top5} \\ 
      \hline
      \hline
      Clean the board $\rightarrow$ takeout pan $\rightarrow$ wash the onion $\rightarrow$ clean the fish $\rightarrow$ cut the onion $\rightarrow$ heat the pan $\rightarrow$ pour oil in pan $\rightarrow$ \textbf{[MASK]} the fish. 
      & fry, cook, boil, wash, clean \\ 
      \hline
      Clean the board $\rightarrow$ takeout pan $\rightarrow$ wash the onion $\rightarrow$ clean the fish $\rightarrow$ cut the onion $\rightarrow$ heat the pan $\rightarrow$ pour oil in pan $\rightarrow$ fry \textbf{[MASK]}. 
      & pan, fish, chicken, it, onion \\ 
      \hline
    \end{tabular}
    \caption{Given a sequence of actions extracted from a video, \textbf{BERT@top5} shows the top5 prediction made by a standard pre-trained BERT for the masked verb and object of the next action.}    
    \label{tab:mlm_ex}
\end{table}
To this end, we propose a simple cross-modal distillation approach, where we distill the knowledge gained by a language model from the text modality of the data into a vision-modality model. 
We build a teacher using a pre-trained language model which already carries general knowledge about action sequences. We adapt this teacher to the action sequences in the video domain by fine-tuning them for the action anticipation task. Then, we train a vision-modality \footnote{The task requires the anticipation model to make inference based on the vision modality (video frames) of the video} student, which is now tasked with both predicting the target action label as well as matching the output probability distribution of the teacher. 

There are two aspects of language models that can be adjusted further for improved distillation. First, while they may contain knowledge about a broad range of action sequences, we can focus them towards specific action sequences in the target dataset. Second, the text modality teacher can be further improved by pretraining on domain-relevant texts (e.g. cooking recipes), to further adapt it to the action sequences in the task domain. 

Our empirical evaluation shows that this cross-modal training yields consistent improvements over a state-of-the-art Anticipative Vision Transformer model~\cite{avt} on two egocentric action anticipation datasets in the cooking domain. Adapting the teacher to the task domain by pretraining on domain relevant texts yields further gains and the gains are stable for different language models. Interestingly, our analysis shows that the language model based teacher can provide gains even when it is not necessarily better than the vision student, suggesting that distillation benefits can also come from the complementary of knowledge, as in the case of the text modality.

In summary we make the following contributions: 
(i) We show that a simple distillation scheme can effectively transfer text-derived knowledge about action sequences (i.e. knowledge external to the video datasets) to a vision-based action anticipation model.
(ii) We show that text-derived knowledge about actions sequences contain complementary information that is useful for the anticipation task, especially for the case where the action label space is large. (iii) Using a strong action anticipation model as a student, we achieve new state-of-the-art results on two benchmark datasets.

\section{Related Work}
There has been a wide range of solutions for action anticipation ranging from hierarchical representaions~\cite{lan2014hierarchical}, unsupervised representation learning~\cite{vondrick2016anticipating}, to encoder-decoder frameworks that decode future actions at different time scales~\cite{rulstm}, and transformers trained on multiple auxiliary tasks~\cite{avt}. However, these only use the vision modality features of the observed video to train the model for the anticipation task. Our work aims to distill text-derived knowledge to improve action anticipation. Here we relate our work to others that have made use of (i) textual knowledge for related tasks, (ii) general knowledge distillation, and (iii) multimodal models which also allow for integration of information from different modalities.
\paragraph{Textual Knowledge for Action Anticipation:} 

Other works have also shown the utility of modeling text-modality. 
\citet{sener2019zero} transfer knowledge in a text-to-text encoder-decoder to a video-to-text encoder-decoder, by substituting the text encoder with the video encoder. However, this relies on projecting the image and text features in a shared space, which requires lots of properly aligned text and its corresponding image. 
~\citet{camporese2021knowledge} model label semantics with a hand engineered deterministic label prior based on the global co-occurrence statistics of the action labels from the overall training data, which can be ineffective in case the underlying joint action distribution is complex.
%
In contrast, our work proposes a different approach to leverage the text in the training data by using language models to learn the complex underlying distribution of action sequences in the video and then distill this knowledge into a vision model to improve their performance.
\paragraph{Cross-modal Knowledge Distillation:} 
%
%
\citet{thoker2019cross} propose learning from RGB videos to recognize actions for another modality.
Others have used cross-modal distillation for video retrieval tasks~\cite{hu2020creating, chen2020imram} and for text-to-speech~\cite{wang2020end}.
Most relevant to ours is a recent system that improves language understanding of text models by transferring the knowledge of a multi-modal teacher trained on a video-text dataset, into a student language model with a text dataset~\cite{tang2021vidlankd} . In contrast, our proposed method for action anticipation transfers knowledge gained by a text-based teacher model into a vision-based student model.

\paragraph{Mutlimodal Models:} Due to the recent prevalence of multimodal data and applications \cite{lin2014microsoft, sharma-etal-2018-conceptual, antol2015vqa, krishna2017visual, ordonez2011im2text, abu2018will, talmor2021multimodalqa, afouras2018deep}, there has been plethora of recent work on multimodal transformers. One commonly used approach used to train these models is to learn a cross-modal representation in a shared space. Examples include learning to align text-image pairs for cross-modal retrieval~\cite{clip, wehrmann2020adaptive}, grounded image representations~\cite{liu2019aligning}, and grounded text representations~\cite{Tan2020VokenizationIL, li2019visualbert}. \citet{hu2021unit} extend the idea for multi-task settings with multiple language-vision based tasks. \citet{tsimpoukelli2021multimodal} adapt a vision model to a frozen large LM to transfer its few-shot capability to a multimodal setting (vision and language). However these methods rely on large-scale image-text aligned datasets for the training the model, which may not always be available, for e.g. \texttt{EGTEA-GAZE+} video dataset has only $10.3$K labelled action sequences. In contrast our distillation approach does not require any image-text alignment for the anticipation task.
\section{Language-to-vision knowledge distillation for action anticipation}
The action anticipation task asks to predict the class label of a future action based on information from an observed video sequence.
In this task setting, the model has access to both, video and annotated action segments (action text) during the train time, but needs to make the inference only using the video sequence.
The input to the prediction model is a sequence of video frames up until time step $t$: $\X = (X_1, X_2, \ldots, X_t)$, and the desired output of the model is the class label $Y$  of the action at time $t + \tau$, where $\tau$ is the anticipation time.

\def\bXi{\X^i}
\def\Xi{X^i}
\def\Yi{Y^i}
\def\bLi{\L^i}
\def\Li{L^i}
\def\ti{t^i}
\def\ki{k^i}

To learn an anticipation model, we assume there is training data of the following form: $\mD = \{ (\bXi, \bLi, \Yi)  \}_{i=1}^{n}$, where $\bXi = (\Xi_1, \ldots, \Xi_{\ti}) $ is the $i^{th}$ training video sequence,  $\Yi$ is the class label of the future action at time $\ti + \tau$, and $\bLi = (\Li_1, \ldots, \Li_{\ki}) $ is the sequence of action label of the action segments in the video sequence $\bXi$. Each human action can span multiple time steps, so so the number of actions $\ki$ might be different from the number of video frames $\ti$. 

Our task is to learn a model $g$ that can predict the future action label based on the vison modality of the video sequence $\bXi$ only. A common approach is to optimize cross entropy loss $\mL$ between the model's predicted label $g(\bXi)$ and the  ground truth label $\Yi$ of each training instances, i.e., to minimize: $\sum_i \mL(g(\bXi), \Yi)$. Although the sequence of action labels $\bLi$ is available in the training data, the semantics associated with these labels is not properly used by the existing methods for training the anticipation model. 

Here we propose to learn a text-based anticipation model $g_{text}$ and use it to supervise the training of the vision-based anticipation model $g$. This training approach utilizes the knowledge from the text domain, which is easier to learn than the vision-based knowledge, given the abundance of event sequences described in text corpora. Hereafter, we will refer to the language-based model as the teacher, and the vision-based model as the student.




\subsection{Overview}
%
\begin{figure*}
    \centering
    \includegraphics[width=0.99\linewidth]{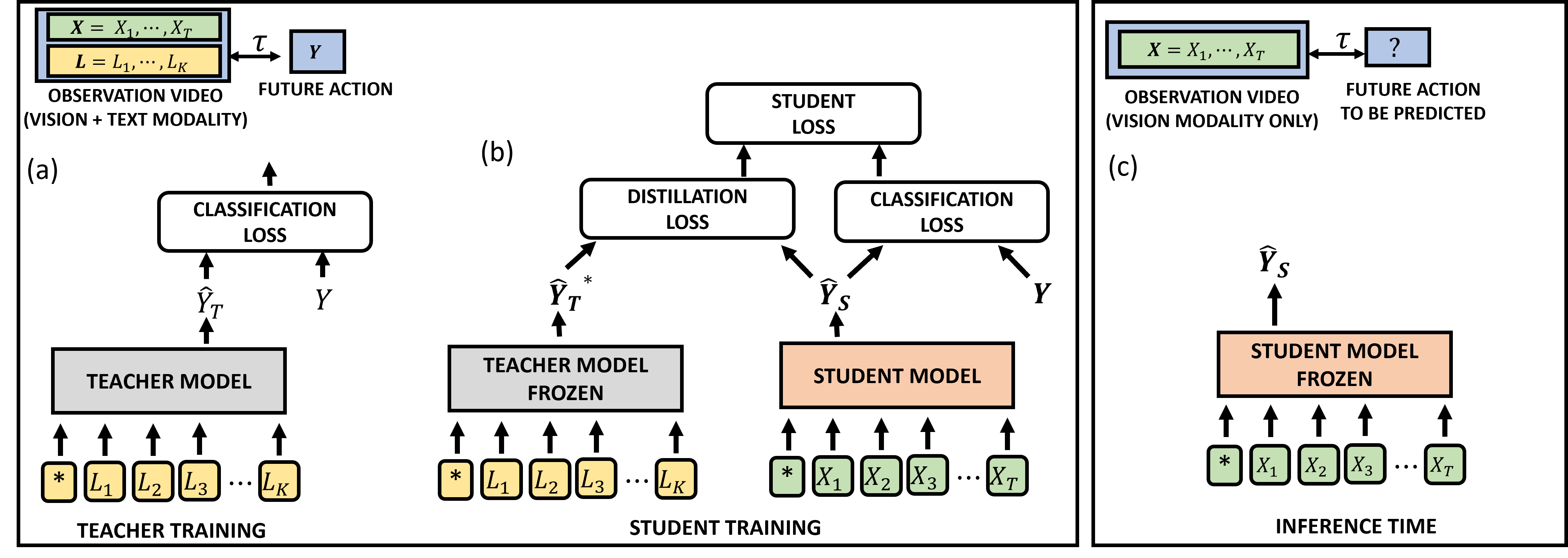}
    \caption{\textbf{METHOD OVERVIEW:} \underline{\textbf{Training}}- The observation video has two set of features, a sequence of $T$ image frames $\mathbf{X}$, and a sequence action labels (\texttt{e.g. cut-onion, peel-onion etc.}) $\mathbf{L}$ corresponding to the $K$ action segments in $\mathbf{X}$. $\mathbf{(a)}$ We train a teacher model to predict $Y$ using the text features $\mathbf{L}$. Then we distill the knowledge gained by the teacher on text features into the student model that operates on vision modality $\mathbf{X}$. For this, $\mathbf{(b)}$, we train a student model on the vision modality feature $\mathbf{X}$ while using the corresponding prediction from the teacher model as a label prior. \underline{\textbf{Inference}}- During the inference or test time, the trained student model is used to predict the future action using only the vision modality of the observed video.
    }
    \label{fig:distill_overview}
    \vspace{-0.5em}
\end{figure*}

The overview of our proposed method is shown in Figure \ref{fig:distill_overview}. 
We augment vision-based anticipation models (students) with knowledge distilled from text-based models (teachers) that have access to knowledge from large scale action sequences. To this end we fine-tune a pre-trained language model on the action sequences in the training data. However, unlike the student, the teacher gets to see the action labels of the input video segment to make its predictions (Figure \ref{fig:distill_overview}a). 
Then, we train a vision-based student that learns from the text-based teacher (Figure \ref{fig:distill_overview}b).

The teacher in our setting is built using a pre-trained language model $g_{txt}$ that has access to broad knowledge about action sequences. We fine-tune it on the target dataset as follows. For each instance, the teacher is given the textual action sequence $\bLi$ in the input video as the input (or conditioning context), which then predicts the anticipated future action $\hat{\Yi}_{txt}$.
The teacher is trained to minimize the loss defined over the predicted and true labels, i.e., 
 to minimize: $\sum_i \mL_{txt}(g_{txt}(\bLi), \Yi)$, where $\mL$ denotes the cross-entropy loss and $\hat{\Yi}_{txt}=g_{txt}(\bLi)$ is the output of the text-based teacher model. 

%
%

We then freeze the \textit{teacher}, and train a vision-based \textit{student} model $g$ that predicts the future action using the vision features $\bXi$. 
The student is trained to minimize the loss $\mathcal{L}_S(\Yi, \hat{Y}^i, \hat{Y}^{i}_{txt})$ such that it's output probability distribution $\hat{Y}^i = g(\bXi)$ matches that of the teacher's output $\hat{\Yi}_{txt}$, in addition to matching the true label $\Yi$.

\subsection{Teacher} \label{subsec:teacher}
The input to the teacher is a sequence of action phrases $\L = (L_1, \cdots, L_k)$ that denotes the sequence of actions observed in the input video segment. The teacher first uses a standard language model $\phi_{LM}$ to produce a vector $f_{txt}$, of the input sequence $\L$. In transformer-based language models, a special token (e.g. \texttt{[CLS]} in \texttt{BERT}) is prepended to the input sequence. The output contextual representation of this special token is used as the final representation of the entire input sequence. 

The teacher uses this $f_{txt}$ vector to predict the output labels using the standard linear transformation ($\mathbf{W}$, b) followed by a softmax layer. In addition we also train the teacher to predict the main verb $Y_{vb}$ and the object $Y_{ob}$ of the action $Y$. These are predicted using separate linear transformations $(\mathbf{W_v}, \mathbf{b_v})$ and $(\mathbf{W_o}, \mathbf{b_o})$, followed by softmax.

%
The full set of predictions for input $\L = (L_1, \cdots, L_k)$ is obtained as:
\begin{align*}
    f_{txt} &= \phi_{LM}(L_1, ..., L_k) \\ 
    \hat{Y}_{txt} &= \mathbf{softmax}(\mathbf{W} f_{txt} + \mathbf{b}) \\ 
    \hat{Y}_{ob} &= \mathbf{softmax}(\mathbf{W_{o}} f_{txt} + \mathbf{b_{o}}) \\ 
    \hat{Y}_{vb} &= \mathbf{softmax}(\mathbf{W_{v}} f_{txt} + \mathbf{b_{v}}) \nonumber 
\end{align*}

\par
To fine-tune the teacher model, we minimize the weighted sum of the cross-entropy loss between the predicted triplet of action, verb and noun and their corresponding ground truth values.

\begin{align}
\mathcal{L}_{txt}(Y,(\hat{Y}_{txt}, \hat{Y}_{ob}, \hat{Y}_{vb})) = \lambda \mL(Y, \hat{Y}_{txt}) \nonumber \\ 
+ \lambda_o \mL(Y_{ob}, \hat{Y}_{ob}) + \lambda_v \mL(Y_{vb}, \hat{Y}_{vb})
\end{align}

\subsubsection{Adapting Teacher using Domain Relevant Texts} 
Pre-trained LMs have been shown to contain a wide variety of knowledge, which we hope to distill into the vision student model. However, there are two aspects about LMs which limit their applicability. First, LMs are general purpose models that cover many domains, but the target video datasets cover specific domains. For example, many action anticipation datasets are built for the cooking domain.  Second, unlike fluent texts that LM's are trained on, the action sequences in the videos are annotated using simpler verb/object constructions. Adapting LMs to these differences can benefit the knowledge distillation. To this end, we make use of domain-relevant texts (for e.g. the recipes in \texttt{Recipe1M} \cite{1Mrecipe} dataset). The recipes are natural language instructions. To mimic the target sequences in the video datasets, we convert these into simpler verb-object constructs, and then use the standard Masked Language Modeling training task. Thus, this allows us to not only make use of generic knowledge about action sequences but also adapt the text-derived knowledge to the target domain.


\subsection{Student}

The student is trained to take the video frames in the video segment $\X=(X_1, ..., X_t)$ as input and predict the future action $Y$ as output. Though the applicability of the proposed distillation method is not restricted to any particular class of student model, we use the recent state-of-the-art Anticipative Vision Transformer (\texttt{AVT}) \cite{avt} as our student model. In \texttt{AVT}, the video to action prediction is done in two stages, first a \emph{backbone} network $\mathcal{B}$ generates the feature representation of the individual frames in $\X$ in a non-contextual manner.
\begin{align}
    & z_1,\ldots, z_t = \mathcal{B}(X_1),\ldots, \mB(X_t) \nonumber &
\end{align}
This is then followed by a transformer based decoder $head$ network $\mathcal{D}$, that generates the contextual representation of the frames by transforming the frame features $z_i$'s in an autoregressive manner.
\begin{align*}
    &f_{v_1}, ..., f_{v_t} = \mathcal{D}(z_1, ..., z_t) \nonumber \\
    &\hat{Y}_{v_j} = \mathbf{softmax}(\mathbf{W}_s f_{v_j} + \mathbf{b}_s) \;\;\;\; \forall j \in \{1, \cdots, t\} \nonumber \\
    &\hat{Y} = \hat{Y}_{v_t} \nonumber
\end{align*}
The feature representations from the $head$ network $f_{v_j}$'s are then used to make predictions for the anticipated action $\hat{Y}_{v_j}$ at time unit $j$. The anticipated action $\hat{Y}$ for the input video $\X$ is simply the predicted label at time unit $t$ i.e. $\hat{Y}_{v_t}$. During training the model is also supervised for two other auxiliary tasks namely future feature prediction and intermediate action prediction (see \cite{avt} for details). We denote this combined training loss function as $\mathcal{L}_{AVT}$.

%
For the teacher to student distillation, we want the AVT's output distribution over action classes $\hat{Y}$ to match the teacher's distribution $\hat{Y}_{txt}$. To this end, we minimize the KL divergence between the teacher prediction $\hat{Y}^{\gamma}_{txt}$ and student predictions $\hat{Y^{\gamma}}$, after smoothing the distributions using a temperature parameter $\gamma$, following the standard distillation technique~\cite{hinton2015distilling}.
\begin{align}
    & \mathcal{L}_S = \mathcal{L}_{AVT} + \lambda_S\cdot\mathcal{D}_{KL}(\hat{Y}^{\gamma}_{txt} \parallel \hat{Y^{\gamma}})
\end{align}
\begin{table}[ht]
\small
\centering
\begin{tabular}{p{0.27\linewidth} p{0.25\linewidth} p{0.15\linewidth} p{0.12\linewidth} }
\hline
\textbf{Dataset} & \textbf{Segments} & \textbf{Classes} & $\mathbf{\tau}$\\
\hline
\hline
\texttt{Epic 55} & $28.6\text{K}\:+9\:\text{K}$ & $2,513$ & $1.0$ sec \\
\hline
\texttt{EGTEA-Gaze+} & $7.3\text{K}\:+3\text{K}$ & 106 & $0.5$ sec \\
\hline
\end{tabular}
\caption{\textbf{Datasets} on which the proposed method is benchmarked. \textbf{Segments} are the number of action segments in the train + test set, \textbf{Classes} are the number of action classes in the dataset, $\mathbf{\tau}$ is the anticipation time.}
\label{Table:dataset}
\end{table}
\section{Experimental Setup}
\subsection{Datasets} \label{subsec:dataset}
\paragraph{1. Anticipation Datasets} We evaluate the proposed method on two different datasets that are summarised in Table.~\ref{Table:dataset}. 
Both the datasets, \texttt{Epic-Kitchen 55} \cite{epic55} and \texttt{EGTEA-GAZE+} \cite{egtea}, are egocentric (first-person) videos of people cooking some recipe. Note the proposed method is broadly applicable to other types of dataset as long as the input video segments in the training set contain action sequence annotations. For the \texttt{Epic-Kitchen 55} dataset, we use the standard train-test split followed in \citet{rulstm}. For the \texttt{EGTEA-GAZE+} dataset, we report performance on the first of the three train-test splits following previous work by~\citet{avt}.
\paragraph{2. Domain-Relevant Dataset}
The teacher can be improved further by adapting its language model (LM) to domain relevant texts. To test the effectiveness of this, we use the \texttt{Recipe1M} dataset \cite{1Mrecipe} to pre-train the LM. The \texttt{Recipe1M} dataset contains one million recipes along with associated images (which are not used in this work). The instructions in a recipe can be seen as a sequence of cooking actions to be performed.
\subsection{Performance Metrics}\label{subsec:perf_met}
For the \texttt{EGTEA-Gaze+}, we report the performance on \texttt{top1 accuracy (\textbf{Acc@1})} and \texttt{class mean recall (\textbf{Rec@1})}-mean recall of the individual classes, as reported by \citet{avt}. 
For the \texttt{Epic-Kitchen 55} dataset, there are a set of action classes that occur only in the train set but not in the test set and vice versa. Existing anticipation methods, including our proposed work does not support zero-shot learning. Therefore \texttt{top5 many-shot class-mean recall (\textbf{MS-Rec@5})}-mean top5 recall of the classes in the many-shot-classes, as mentioned in \citet{furnari2018leveraging}, is our primary metric for model evaluation.
\subsection{Implementation Details} \label{subsec: implementation}
\paragraph{1. Teacher Training:} The teacher model is a classification layer on top a pre-trained language model. For the main set of experiments we used \textbf{\texttt{AlBERT}}~\cite{albert} as the base language model. Our choice here is motivated by two main factors: (i) the pre-training task for \textbf{\texttt{AlBERT}} focuses on modeling the inter-sentence coherence which is important when modeling the sequence of disparate action phrases (ii) it enables faster training of deeper models.
For the \texttt{EGTEA-GAZE+} dataset, we trained the model for 4 epochs by minimizing the weighted cross-entropy loss (inversely weighted by the relative class frequency) due to the high degree of class imbalance in the dataset ($\sim1:24$). For the \texttt{EPIC-Kitchen-55} dataset, the model was trained for 8 epochs using regular cross-entropy loss instead of weighted cross-entropy as a lot of classes in the test label set are not present in the train-set, and vice versa.

The classification head is a single linear layer ($\mathbf{W} \cdot x + \mathbf{b}$) that projects the feature representation of the input action sequence into the label space of the target dataset.
For optimizing on both the datasets, we used the AdamW  (\cite{adamw}) optimizer, with a learning rate of $10^{-5}$ and weight decay of $10^{-7}$. The context window for the Epic-Kitchen was set to 5 action segments whereas for the \texttt{EGTEA-GAZE+} it was set to 15 action segments. The teacher training was performed on two Nvidia RTX Titan-X GPUs. The teacher training for the \texttt{EGTEA-GAZE+} takes about 2-4 hours depending on the LM base whereas the \texttt{EPIC-Kitchen-55} takes about 3-5 hours to train.
\paragraph{2. Teacher Pre-training:}
We first parse each instruction in the \texttt{Recipe1M} dataset into a sequence event tuples of the form (subject, verb, object) using an open information extraction system~\cite{openie} made available by AllenNLP~\cite{allennlp}. To match the action label structure we see in the video datasets, we represent each instruction using the sequence of action, i.e. <verb, object> part of the event. The actions in the action sequence are sorted by the discourse order of their corresponding verb in the instruction. The language model is pretrained on these (verb, object) sequences using the standard masked language modeling objective \cite{bert}, where some token in the sequence is masked at random and the model is tasked with predicting the masked token.

For pre-training, the language models were trained on the \texttt{Recipe1M} dataset for 200K steps with a batch size of 16. The optimizer used was AdamW \cite{adamw}, with a learning rate of $10^{-5}$ and weight decay of $10^{-7}$. LM pre-training was performed on a single Nvidia A100 GPU with the training time varying from 12 hrs for the smallest model (\texttt{\textbf{DistilBERT}}) to 24 hrs for \texttt{\textbf{BERT, RoBERTa}}, and \texttt{\textbf{AlBERT}}.
\paragraph{3. Student Training:}
For the student training, all the hyperparameters and initial conditions (parameter initialization) are exactly identical to the ones used to train the \textbf{\texttt{AVT}}~\cite{avt} baseline model. So any change in the performance from the baseline is the result of adding the knowledge distillation. 
The distillation loss coefficient $\lambda_S$, for the \texttt{EGTEA-GAZE+} dataset was set to $150$, and $20$ for \texttt{EPIC-Kitchen 55}.
\paragraph{4. Top-K logit distillation:}
The label space of \texttt{EPIC-Kitchen 55} has $2,513$ classes, out of which only 31\% of the classes in the training data are present in the test data. This leads to the teacher model assign relatively low probability values to many classes, which may not be reliable signals for distillation. 
Therefore, instead of matching the probability distribution over all the action classes, we only match the relative probability distribution of the top-50 classes with the highest teacher probabilities. For this, we consider the classes corresponding to the top-50 logits from the teacher prediction, normalize them, and only minimized the KL-Divergence between them and their corresponding logits of the student prediction.
The student training was performed on either Nvidia Tesla V100 GPU and the training time was $\sim24$ hrs for \texttt{EGTEA-GAZE+} and $\sim6$ hrs for the \texttt{EPIC-55 dataset}.
\section{Results and Analysis}

We present the results of text to video knowledge distillation on the \textbf{\texttt{AVT}} \cite{avt} model as the student. AVT is the state-of-the-art model for action anticipation on the \texttt{EGTEA-GAZE+} and \texttt{EPIC-Kitchen 55} datasets on all performance metrics.
\paragraph{}\label{par:avt_variant}For each of these datasets, we consider the AVT variants with the best performance as our baseline and student model.
For the \texttt{EGTEA-GAZE+} dataset, we consider \textbf{\texttt{AVT-h + AVT-b}} in \cite{avt} as our baseline model. 
Similarly for the \texttt{EPIC-Kitchen 55} dataset, we consider, \textbf{\texttt{AVT-h +  irCSN152}} in \cite{avt} as our baseline model.
\footnote{\label{ftn:model}$^\clubsuit$\textbf{\texttt{AVT}} variants used for the \texttt{EGTEA-GAZE+} and \texttt{EPIC-Kitchen 55} baselines are \textbf{\texttt{AVT-1}} and \textbf{\texttt{AVT-2}}.}
Throughout this section, we refer to \textbf{\texttt{AVT-h + AVT-b}} and \textbf{\texttt{AVT-h +  irCSN152}} as \textbf{\texttt{AVT-1}} and \textbf{\texttt{AVT-2}} respectively.
The baseline models distilled with LM based teacher is denoted as \textbf{\texttt{AVT-1(or \textbf{2) + LM Distillation}}} and in case teacher LM is pre-trained on the recipe domain text, the resulting model is referred to as \textbf{\texttt{AVT-1(or \textbf{2) + RcpLM Distillation}}}. We tried to reproduce the \textbf{\texttt{AVT}} model to use as our student and obtain stronger results than the published version (see Table~\ref{Table:result3}), on all but one metric. We use this stronger implementation as our baseline and our student model.

\subsection{Does Knowledge distillation from Language Models help ?}
Table~\ref{Table:result3} shows the result of training the state-of-the-art baseline model \texttt{\textbf{AVT}}, with and without the text to vision knowledge distillation, for the \texttt{EGTEA-GAZE+} and \texttt{EPIC-Kitchen 55} dataset.
We can observe that applying text to vision knowledge distillation to the \textbf{\texttt{AVT}} leads to performance gains on both the datasets. For \texttt{EGTEA-GAZE+}, adding knowledge distillation leads to $2.1\%$ and $2\%$ relative percentage improvement over \textbf{\texttt{AVT-1}} on the \textbf{Acc@1} and \textbf{Rec@1} metrics respectively.
For the \texttt{EPIC-Kitchen 55} dataset, knowledge distillation leads to a relative performance gain of $3.5\%$ over \textbf{\texttt{AVT-2}} on \texttt{\textbf{MS-Rec@5}} metric. 
\subsection{Does domain-adaptive pre-training of teacher improves the task performance ?}
To analyze the effect domain adaptive pre-training on the task, we pre-train the teacher LM on the \texttt{Recipe1M} dataset through the MLM task. The pre-trained model was then finetuned on the task-specific video dataset for the anticipation task. As seen in Table~\ref{Table:result3}, the performance gain of the teacher directly translates to the performance gain of the student. 
For \texttt{EGTEA-GAZE+} dataset, pre-training teacher leads to $3.9\%$ and $3.4\%$ relative improvement over the \textbf{\texttt{AVT-1}} on \textbf{\texttt{Acc@1}} and \textbf{\texttt{Rec@1}} metric compared to $2.1\%$ and $2\%$ relative improvement when not pretraining the teacher. 
For the \texttt{EPIC-Kitchen 55} dataset, teacher pre-training leads to a relative improvement of $7.2\%$ on \texttt{\textbf{MS-Rec@5}} metric compared to only $3.5\%$ when not pre-training the teacher.

\begin{table}[!h]
\small
\centering
{\renewcommand{\arraystretch}{1.4}
\begin{tabular}{m{0.35\linewidth} | m{0.12\linewidth} | m{0.12\linewidth} | m{0.2\linewidth}}\Xhline{4\arrayrulewidth}
 \multirow{ 2}{*}{\textbf{Model}} & 
 \multicolumn{2}{c|}{\small{\textbf{\texttt{EGTEA-GAZE+}}}} &
 \small{\textbf{\texttt{EPIC-55}}} \\\cline{2-4}
&  \textbf{\small{\makecell[l]{\texttt{Acc@1}}}} & 
 \textbf{\small{\makecell[l]{\texttt{Rec@1}}}} & 
 \textbf{\small{\makecell[l]{\texttt{MS-Rec@5}}}} \\ \hline \hline
\small{\makecell[l]{\textbf{\texttt{AVT}}}} \cite{avt} \texttt{\textbf{-published}} & $43.0$ &  $35.5$  & $13.6$ \\ \hline
\small{\makecell[l]{\textbf{\texttt{AVT}}}} \cite{avt} \texttt{\textbf{-reproduced}} & $43.52$ &  $34.87$  & $15.25$ \\ \Xhline{4\arrayrulewidth}
 \small{\makecell[l]{\textbf{\texttt{AVT + LM}}\\ \textbf{\texttt{Distillation}}}} & $44.41$ &  $35.54$ & ${15.79}$ \\ \hline
 \small{\makecell[l]{\textbf{\texttt{AVT + RcpLM}}\\ \textbf{\texttt{Distillation}}}} & $\mathbf{45.2}$ & $\mathbf{36.1}$ & $\mathbf{16.36}$ \\ \Xhline{4\arrayrulewidth}
\end{tabular}}
\caption{\textbf{Effect of knowledge distillation:} Distilling knowledge from teacher (\texttt{\textbf{ALBERT}} LM) trained on text-modality of the video data, into vision based student model leads to student performance gain. Pre-training the teacher on domain relevant text before task specific finetuning leads to further performance improvement.\hyperref[ftn:model]{$^\clubsuit$}
}
\label{Table:result3}
\end{table}

\subsection{How sensitive is the distillation to the choice of Language Model ?}
In order to analyze the sensitivity of the distillation scheme towards the choice of the language model, we also trained multiple teachers with different pre-trained LMs. The result of using different teachers for the anticipation task is specified in Table~\ref{Table:result5}.
From the table, we can observe that all the teacher distilled models perform better than the baseline \texttt{\textbf{AVT}} on all the metrics for both the datasets.
This indicates that the text modality has some information that complementary to the vision modality that if properly exploited can lead to improved performance for the anticipation task.

\begin{table}[t!]
\centering
{\renewcommand{\arraystretch}{1.4}
\begin{tabular}{m{0.37\linewidth} | m{0.12\linewidth} | m{0.12\linewidth} | m{0.19\linewidth}}
\Xhline{4\arrayrulewidth}
 \multirow{ 2}{*}{\textbf{Model}} & 
 \multicolumn{2}{c|}{\small{\textbf{\texttt{EGTEA-GAZE+}}}} &
 \small{\textbf{\texttt{EPIC-55}}} \\\cline{2-4}
&  \textbf{\small{\makecell[l]{\texttt{Acc@1}}}} & 
 \textbf{\small{\makecell[l]{\texttt{Rec@1}}}} & 
 \textbf{\small{\makecell[l]{\texttt{MS-Rec@5}}}} \\ \hline \hline
     \small{\makecell[l]{\textbf{\texttt{AVT}}}} \cite{avt} 
     &  $43.52$ & $34.87$  & $15.25$ \\ \Xhline{4\arrayrulewidth}
     \small{\makecell[l]{\textbf{\texttt{ + Rcp-ALBERT}}\\ \textbf{\texttt{Distillation}}}}     
     & $45.2$  & $36.1$ & $\mathbf{16.36}$ \\ \hline
     \small{\makecell[l]{\textbf{\texttt{ + Rcp-BERT}}\\ \textbf{\texttt{Distillation}}}}      
     & $44.81$  & $35.57$  & ${15.98}$ \\ \hline
     \small{\makecell[l]{\textbf{\texttt{ + Rcp-RoBERTa}}\\ \textbf{\texttt{Distillation}}}}  
     & $\mathbf{45.5}$  & $\mathbf{36.53}$ & ${15.97}$ \\ \hline
     \small{\makecell[l]{\textbf{\texttt{ + Rcp-ELECTRA}}\\ \textbf{\texttt{Distillation}}}}  
     & $45.2$  & $35.58$  & ${15.34}$   \\ \hline
     \small{\makecell[l]{\textbf{\texttt{+ Rcp-DistillBERT}}\\ \textbf{\texttt{Distillation}}}} 
     & $44.86$  & $35.64$ & ${16.23}$  \\
     \Xhline{4\arrayrulewidth}
\end{tabular}}
\caption{\textbf{Effect of the choice of teacher LM} on the distillation performance. Each of the pre-trained LMs that we tested as a teacher, showed performance gain over the baseline \textbf{\texttt{AVT}} model for both the datasets.\hyperref[ftn:model]{$^\clubsuit$}}
\label{Table:result5}
\end{table}
     
\subsection{Should the teacher be always better than the student ?}

To understand the impact of the quality of the teachers, we measured the performance of the teacher models by themselves on the anticipation task as show in Table~\ref{Table:result7}. 
For the \texttt{EPIC-Kitchen 55} dataset the teacher performance is much better than the video-only baseline, whereas, for the \texttt{EGTEA-GAZE+} dataset, the baseline vision model's performance is much better than any of the teachers. Despite this, the performance gain due to distillation is greater for the \texttt{EGTEA-GAZE+} dataset compared to the \texttt{EPIC-Kitchen 55} dataset, as seen in Table~\ref{Table:result3}.
This suggests that what matters more for distillation in this case is the complementary information gained from the text modality that is not already present in the vision modality.
\begin{table}[!h]
\centering
{\renewcommand{\arraystretch}{1.4}
\begin{tabular}{m{0.37\linewidth} | m{0.12\linewidth} | m{0.12\linewidth} | m{0.19\linewidth}}
\Xhline{4\arrayrulewidth}
  \multirow{ 2}{*}{\textbf{Model}} & 
 \multicolumn{2}{c|}{\small{\textbf{\texttt{EGTEA-GAZE+}}}} &
 \small{\textbf{\texttt{EPIC-55}}} \\\cline{2-4}
&  \textbf{\small{\makecell[l]{\texttt{Acc@1}}}} & 
 \textbf{\small{\makecell[l]{\texttt{Rec@1}}}} & 
 \textbf{\small{\makecell[l]{\texttt{MS-Rec@5}}}} \\ \hline \hline
\small{\makecell[l]{\textbf{\texttt{AVT}}}} \cite{avt}        
 &  $\mathbf{43.52}$  & $\mathbf{34.87}$           & $15.25$ \\ \Xhline{4\arrayrulewidth}
 \small{\makecell[l]{\textbf{\texttt{Rcp-ALBERT}} \\ \textbf{\texttt{Teacher}}}}          
 & $21.66$            & $22.63$         & ${21.78}$ \\ \hline
 \small{\makecell[l]{\textbf{\texttt{Rcp-BERT}} \\ \textbf{\texttt{Teacher}}}}
& $22.05$            & $23.39$        & ${21.43}$ \\ \hline
 \small{\makecell[l]{\textbf{\texttt{Rcp-RoBERTa}} \\ \textbf{\texttt{Teacher}}}}
& ${19.98}$          & ${21.58}$       & $\mathbf{22.41}$ \\ \hline
\small{\makecell[l]{\textbf{\texttt{Rcp-ELECTRA}} \\ \textbf{\texttt{Teacher}}}}
 & $21.46$            & $23.71$        & ${15.19}$   \\ \hline
 \small{\makecell[l]{\textbf{\texttt{Rcp-DistillBERT}} \\ \textbf{\texttt{Teacher}}}}
 & $21.86$            & $22.58$         & ${21.56}$ \\ \Xhline{4\arrayrulewidth}
\end{tabular}}
\caption{\textbf{Teacher performance} on the anticipation task. For the \texttt{EGTEA-GAZE+} dataset, the teacher performance is much lower than the video only \textbf{\texttt{AVT}} model, where as for the \texttt{EPIC-Kitchen 55} dataset, the teacher performance is much better than the video-only \textbf{\texttt{AVT}} model.\hyperref[ftn:model]{$^\clubsuit$}}
\label{Table:result7}
\end{table}
\section{Conclusions}
\label{sec:conclusions}
Action anticipation is a challenging problem that requires training large capacity video models. In this work, we showed how the textual modality of the input videos, which is often ignored in training, can be leveraged to improve the performance of the video models. In particular, we can exploit the large scale knowledge acquired by pre-trained language models to build a text-modality teacher that can provide useful complementary information about the action sequences to a vision modality student. This cross-modal distillation strategy yields consistent gains achieving new state-of-the-art results on multiple datasets. Last, the gap between the performance of the teacher and the student models for domains with large label space suggests that there is still room for improvement with better distillation techniques.
\section{Limitations}
\label{sec:limitations}

    \par Real life scenarios have a large space of human actions which cannot be exhaustively covered by manually annotated training data. As such it is important to have models with zero-shot anticipation capabilities to predict unseen actions. This work did not explore zero-shot settings but we believe text-to-video distillation holds promise given the recent successes of language models in zero-shot tasks.
    
    \par In this work we have shown the capability of text based language models for action anticipation, especially when the action space is very large and sparse. Though this work is intended to be a proof of concept for leveraging text based model for improving video based action anticipation, there is still a large performance gap between the a text based language model and vision modality model. This performance gap indicates fruitful research avenues in text to vision knowledge distillation for action anticipation task.
\section{Ethical Considerations}
\label{ethics}
Anticipation future action based on videos is an important for many applications such as assistive technologies, augmented reality etc. 
Our work demonstrates that knowledge derived from text sources can be used to further improve the performance of video based action anticipation model. 
Even though our proposed work is able to improve the current state-of-art numbers on the standard benchmark datasets, the absolute performance is still low, especially in the case where the action space is very large. 
As such we would recommend to carefully analyze the cost of erroneous prediction before deploying the system for real world application.
    
Since the proposed method involves distilling the knowledge gained by pre-trained language model from text sources into a vision based model for action anticipation, this can also transfer the biases that these languages models can learn from the training text. As such data on which these text-based teacher models are trained should be analyzed for potential biases before deploying the proposed system for actual application. Analysis of bias propagation during knowledge distillation and devising bias reduction techniques are some potential extension of this work that we are highly interested in.
\section{Acknowledgement}
This material is based on research that is supported in part by the Air Force Research Laboratory (AFRL), DARPA, for the KAIROS program under agreement number FA8750-19-2-1003 and in part by the National Science Foundation under the award IIS \#2007290.

\bibliography{anthology,custom}
\bibliographystyle{acl_natbib}

\end{document}